\documentclass[lettersize,journal]{IEEEtran}
\usepackage{amsmath,amsfonts}
\usepackage{algorithmic}
\usepackage{algorithm}
\usepackage{array}
\usepackage{graphicx}
\usepackage[caption=false,labelfont=sf,textfont=sf]{subfig}
\usepackage{textcomp}
\usepackage{stfloats}
\usepackage{url}
\usepackage{verbatim}
\usepackage{graphicx}
\usepackage{cite}
\usepackage{placeins} 
\usepackage{amssymb}  
\usepackage{booktabs}  
\usepackage[colorlinks=true, pdfborder={0 0 0}]{hyperref} % 
\hypersetup{
    linkcolor=blue,    
    citecolor=blue,   
    urlcolor=black       
}
\usepackage{orcidlink} 
\usepackage{orcidlink} 
\usepackage{wasysym}
\usepackage{graphicx}

\begin{document}

\title{Convolutional Feature Enhancement and Attention Fusion BiFPN\\ for Ship Detection in SAR Images}

\author{Liangjie Meng\textsuperscript{\orcidlink{0009-0002-4865-8779}},~Danxia Li\textsuperscript{\orcidlink{0000-0001-6285-9038}},~Jinrong He\textsuperscript{\orcidlink{0000-0003-4040-4766}},~Lili Ma\textsuperscript{\orcidlink{0009-0009-7738-0201}}, and~Zhixin Li\textsuperscript{\orcidlink{0009-0001-9940-1503}}
        % <-this % stops a space
%\thanks{This paper was produced by the IEEE Publication Technology Group. They are in Piscataway, NJ.}% <-this % stops a space
\thanks{This document is the results of the research project funded by the Natural Science Basic Research Program of Shaanxi (Grant No. 2024JCYBMS576), the National Natural Science Foundation of China (Grant No. 62366053). 
 
Liangjie Meng (email: mengliangjie@yau.edu.cn), Danxia Li (email: ldx@yau.edu.cn), Lili Ma (email: lilimalili0724@gmail.com), Jinrong He (email: hejinrong@yau.edu.cn), and Zhixin Li (email: lizhixin@yau.edu.cn) are with the School of Mathematics and Computer Science, Yan'an University (Corresponding author: Danxia Li.) 
}}
\markboth{Journal of \LaTeX\ Class Files,~Vol.~14, No.~8, August~2021}%
{Shell \MakeLowercase{\textit{et al.}}: A Sample Article Using IEEEtran.cls for IEEE Journals}


\maketitle
\begin{abstract}

Synthetic Aperture Radar (SAR) enables submeter-resolution imaging and all-weather monitoring via active microwave and advanced signal processing. 
Currently, SAR has found extensive applications in critical maritime domains such as ship detection. However, SAR ship detection faces several challenges, including significant scale variations among ships, the presence of small offshore vessels mixed with noise, and complex backgrounds for large nearshore ships. To address these issues, 
this paper proposes a novel feature enhancement and fusion framework named C-AFBiFPN.
C-AFBiFPN constructs a Convolutional Feature Enhancement (CFE) module following the backbone network, aiming to enrich feature representation and enhance the ability to capture and represent local details and contextual information.
Furthermore, C-AFBiFPN innovatively integrates BiFormer attention within the fusion strategy of BiFPN, creating the AFBiFPN network.
AFBiFPN improves the global modeling capability of cross-scale feature fusion and can adaptively focus on critical feature regions.
The experimental results on SAR Ship Detection Dataset (SSDD) indicate that the proposed approach substantially enhances detection accuracy for small targets, robustness against occlusions, and adaptability to multi-scale features.
The code is available at \url{https://github.com/mlj666219/C-AFBiFPN/tree/master}. 
\end{abstract}


\section{Introduction}

\IEEEPARstart{S}{ynthetic} Aperture Radar (SAR) is a microwave-based active imaging technology widely used in remote sensing. It operates by emitting electromagnetic waves from radar systems mounted on satellites or aircraft, which then receive the reflected signals from the Earth's surface~\cite{1}. By leveraging the synthetic aperture principle and advanced signal processing algorithms, SAR can produce high-resolution two-dimensional images, regardless of weather or lighting conditions~\cite{2}. This all-weather, day-and-night capability makes SAR especially effective for monitoring dynamic and complex environments such as oceans and polar regions. These imaging characteristics make SAR a core technology for both nearshore and offshore ship detection, with broad applications in illegal fishing surveillance, maritime traffic monitoring, and search and rescue operations~\cite{3}.

Traditional SAR ship detection techniques can be grouped into thresholding~\cite{4}, statistical modelling~\cite{5}, and hand-crafted feature machine learning~\cite{6}. Threshold methods extract targets by applying global or adaptive grey-level limits~\cite{7}, but they are highly vulnerable to speckle and near-shore clutter. Statistical approaches such as maximum-likelihood estimation (MLE) assume specific pixel distributions~\cite{8}; their accuracy collapses when real sea clutter departs from these models. Feature learning schemes draw on texture, edge or intensity descriptors~\cite{9}, yet they rely heavily on manual design and generalise poorly across scales and imaging conditions. Taken together, these classical pipelines struggle to cope with simultaneous scale variation, complex backgrounds and noise, limiting their robustness in practical SAR scenarios.

\begin{figure}[t]
    \centering
    \includegraphics[width=\linewidth]{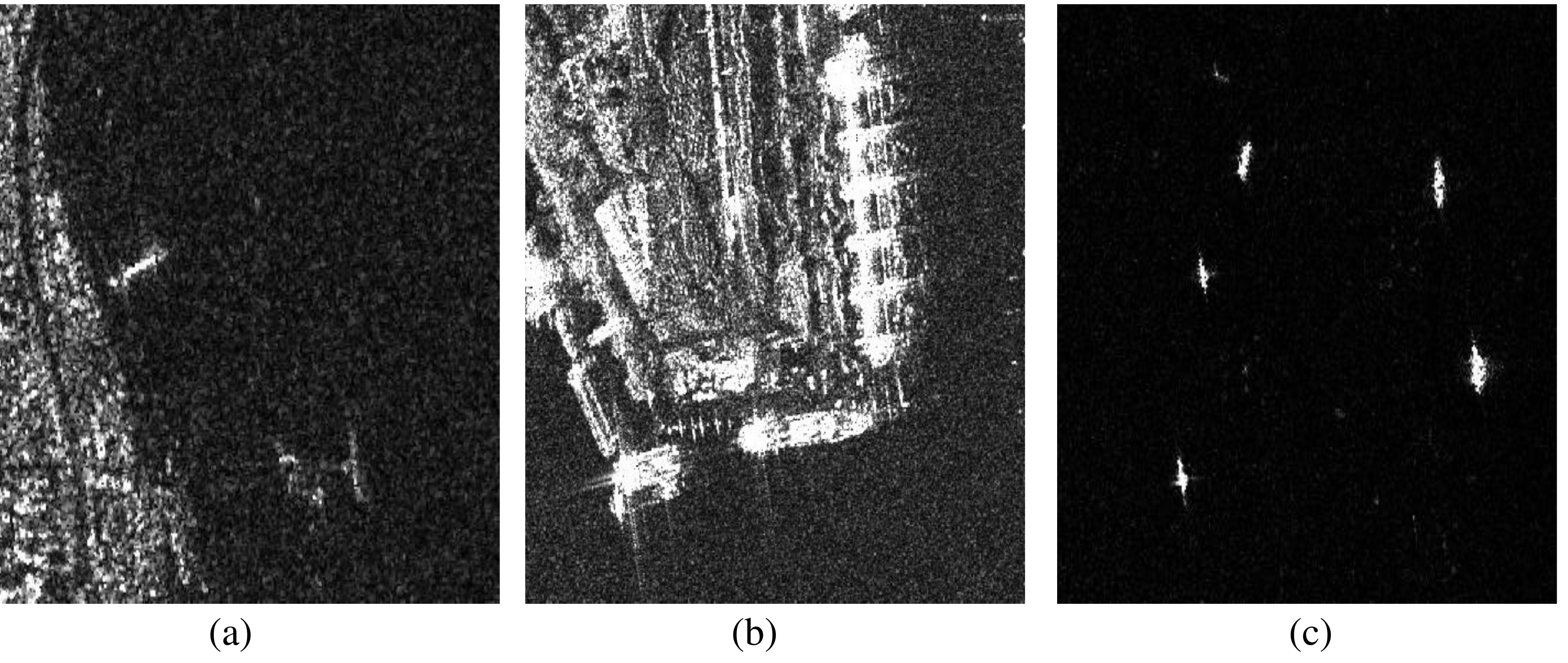}  
    \caption{SAR ship images from the SSDD dataset. (a) and (b) show ships in complex near shore backgrounds, but with significant scale variations. (c) shows multiple small ships in an off-shore scenario, which are affected by the scattered noise on the sea surface.}
    \label{fig:1}
\end{figure}

With the emergence of deep learning, convolutional neural networks have demonstrated superior performance in SAR ship detection tasks due to their ability to learn hierarchical feature representations~\cite{10,11}. Deep learning-based object detection methods are generally categorized into single-stage and two-stage approaches. Single-stage detectors, such as YOLO~\cite{12} and SSD~\cite{13}, are known for their real-time inference and computational efficiency. For instance, Yang et al.~\cite{14} proposed a lightweight YOLO-based method for high-speed ship detection, while Zhou et al.~\cite{15} enhanced YOLOv5 by introducing dynamic scale matching, achieving an AP of 0.515 in the SAR Ship Detection Dataset (SSDD). Despite their speed advantage, single-stage methods often lag behind in detection accuracy compared to two-stage detectors.

Two-stage methods, such as R-CNN, Fast R-CNN~\cite{16}, and Faster R-CNN~\cite{17}, provide higher detection accuracy by first generating candidate regions and then refining the predictions. Faster R-CNN, in particular, integrates a trainable Region Proposal Network (RPN), significantly improving detection precision without excessive computational cost. Zhou et al.~\cite{18} applied Faster R-CNN with deep features for large ship detection, achieving an AP$_L$ of 0.533, while Liu et al.~\cite{18} enhanced its backbone to allow multi-level feature fusion, leading to a 21\% AP improvement over the baseline.

However, as shown in Fig.~\ref{fig:1}, SAR ship detection remains challenging due to the complex background and significant
scale variations that affect ships near shore, as well as the interference of sea surface noise on small ships in offshore scenarios. Existing detectors like Faster R-CNN, which rely solely on single-scale deep features or separate scale-wise predictions, often struggle with small and medium-sized ships in cluttered scenes. To address these issues, Feature Pyramid Networks (FPN) were introduced to aggregate features across multiple levels. Liu et al.~\cite{19} adopted Faster R-CNN with FPN and achieved a 42\% improvement in AP by leveraging multi-scale fusion. But, standard FPN inadequately utilize low-level features, limiting their effectiveness in small object detection under noisy SAR conditions. Tan et al.~\cite{20} proposed BiFPN, which iteratively fuses features via bidirectional, learnable-weighted paths and thus greatly improves multi-scale feature aggregation efficiency. Cao et al.~\cite{21} presented FBFPN, which first recalibrates the channels of shallow and deep layers to alleviate information imbalance, strengthening the complementarity between semantics and texture. Wang et al.~\cite{22} introduced OD-FPN, which uses OctaveConv to split high- and low-frequency features and then integrates them in a lightweight manner, reducing computational cost. 
Despite progress in feature fusion and multi-scale detection, these methods still face limitations in complex scenarios. Insufficient modeling of local details and context results in poor performance for small objects and occlusions. Their fusion strategies also lack flexibility and adaptability, and their weak global modeling restricts generalization and accuracy.

To address the challenges of scale variability, sea-surface interference, and blurred contours near land-sea boundaries in SAR imagery, we propose a novel feature enhancement and fusion framework named \textbf{C-AFBiFPN}. The key contributions of this paper are as follows.

\begin{itemize}
  
    \item We design a Convolutional Feature Enhancement (\textbf{CFE}) module after the backbone; by synergistically integrating multi-branch, deformable, and dilated convolutions, the CFE enriches feature diversity, expands the receptive field, and strengthens both local detail and contextual information in the backbone representations.
 
    \item We improve the intermediate fusion strategy of BiFPN by integrating BiFormer Attention (BA), leading to a new attention-based fusion network, \textbf{AFBiFPN}. 
    AFBiFPN enhances the global modeling capability of cross-scale feature fusion and can adaptively focus on critical feature regions. This not only effectively improves the performance of small object detection, occlusion resistance, and multi-scale adaptability, but also maintains high computational efficiency.
 
    \item We propose the complete \textbf{C-AFBiFPN} framework within the Faster R-CNN pipeline and conduct extensive experiments in the SSDD. Compared with 25 state-of-the-art detectors, our method achieves the highest performance in AP, AP$_S$, AP$_M$, and AP$_L$. Ablation studies further verify the individual contributions of CFE and AFBiFPN to overall detection accuracy.
\end{itemize}

\section{Methods}

\subsection{Overall architecture of C-AFBiFPN}
\begin{figure}[t]
    \centering
    \includegraphics[width=0.47\textwidth, height=0.26\textheight]{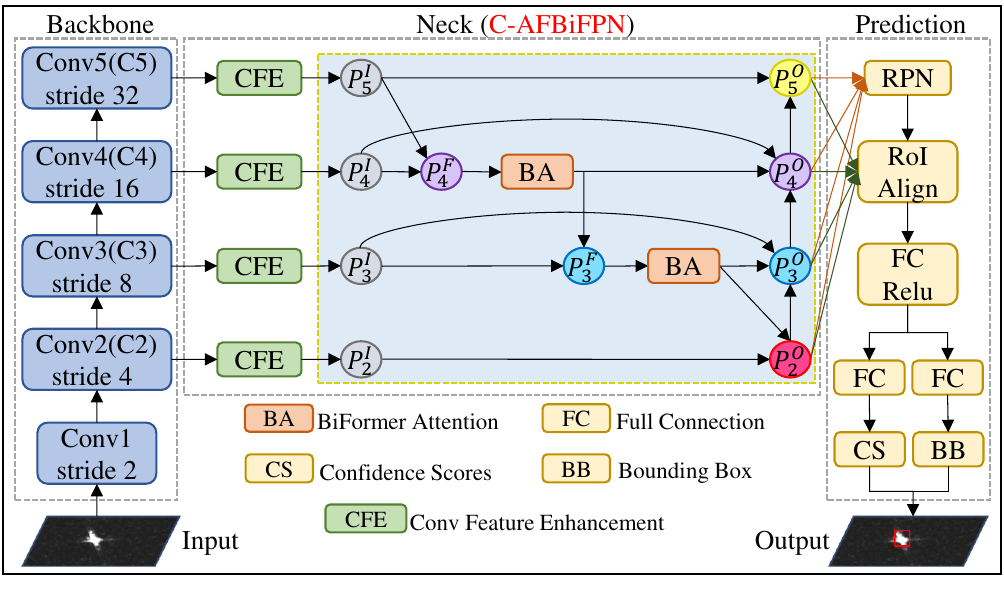}  % Scale the image to 80% of the text width
    \caption{Overall architecture of C-AFBiFPN.}
    \label{fig:2}
\end{figure}

As shown in Fig.~\ref{fig:2}, the C-AFBiFPN framework we proposed consists of two components: CFE and AFBiFPN. 
We employ ImageNet-pretrained ResNet as the backbone network.

The features of backbone levels 2 through 5, denoted as ${C2, C3, C4, C5}$ are passed to the CFE module. By delivering targeted feature compensation and a second-stage receptive-field expansion, CFE supplies richer semantic feature maps for the multi-scale fusion stage in the C-AFBiFPN framework.

After enhancement by the CFE module, the feature maps of each level are labeled ${P_{2}^I, P_{3}^I, P_{4}^I, P_{5}^I}$ and used as input for the AFBiFPN network. The top-down fused feature maps ${P_{4}^F, P_{3}^F}$ produced by AFBiFPN are then refined with BA. BA is a bi-level routing dynamic sparse attention mechanism that enriches ship-detail representations in complex scenes without a marked increase in computational cost. Finally, the BA-refined feature maps are fused with lower-level features, strengthening scale-specific cues for ships, and improving multiscale ship detection performance under complex conditions.

The feature maps output by the C-AFBiFPN network are denoted as ${P_{2}^O, P_{3}^O, P_{4}^O, P_{5}^O}$. These multi-scale feature maps are then passed to RPN, which produces anchor boxes and supplies high-quality proposals for the subsequent object-detection stage.
Next, RoI Align is used to align the candidate region coordinates with the shared feature maps, facilitating processing by the following fully connected layers. 
Finally, the features are passed into the fully connected layers for classification and regression, ultimately yielding the predicted output.

\subsection{Convolutional Feature Enhancement (CFE)} 
The early layers of the current backbone network fail to capture adequate contextual cues for small ships, whereas the deeper layers tend to lose crucial fine-grained details of the ships.
\begin{figure}[ht]
    \centering
    \includegraphics[width=\linewidth]{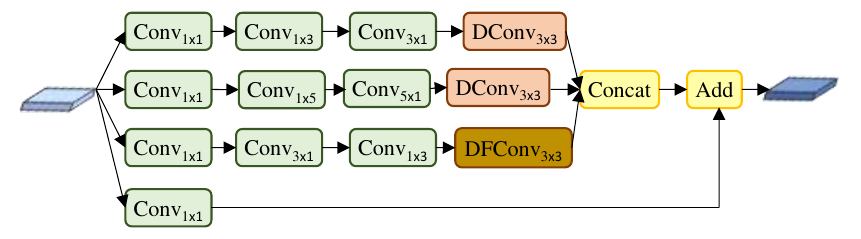}  
    \caption{Convolutional Feature Enhancement (CFE).}
    \label{fig:3}
\end{figure}

To this end, we design a CFE module that simultaneously enriches feature diversity and enlarges the receptive field. As depicted in Fig.~\ref{fig:3}, the CFE adopts a parallel, multi-branch architecture; each branch employs kernels of different sizes and shapes to capture complementary semantic cues, thereby preserving fine-grained textures while injecting higher-level discriminative information. The branch outputs are subsequently merged with the original input via a residual addition, which further strengthens the overall representation. 
For receptive-field enlargement, the CFE integrally fuses dilated and deformable convolutions. This hybrid design adaptively aligns the convolutional sampling grid with object geometries while stretching the spatial coverage, effectively broadening the perceptual range of the network. As a result, the module supplies downstream stages with richer and more comprehensive spatial context.

The mathematical expressions of CFE can be written as follows:
\begin{equation}
\mathrm{B_{1}=DConv_{3\times3}\langle Conv_{3\times1}\{Conv_{1\times3}[Conv_{1\times1}(F)]\}\rangle},
\end{equation}
\begin{equation}
    \mathrm{B_{2}=DConv_{3\times3}\langle Conv_{5\times1}\{Conv_{1\times5}[Conv_{1\times1}(F)]\}\rangle},
\end{equation}
\begin{equation}
    \mathrm{B_{3}=DFConv_{3\times3}\langle Conv_{1\times3}\{Conv_{3\times1}[Conv_{1\times1}(F)]\}\rangle},
\end{equation}
\begin{equation}
    \mathrm{Y=Concat(B_{1},B_{2},B_{3})+Conv_{1\times1}(F)},
\end{equation}
where \( \mathrm{Conv}_{1 \times 1} \), \( \mathrm{Conv}_{1 \times 3} \), \( \mathrm{Conv}_{3 \times 1} \), \( \mathrm{Conv}_{1 \times 5} \), \( \mathrm{Conv}_{5 \times 1} \) and \( \mathrm{Conv}_{3 \times 3} \) represent standard convolution operations with kernel dimensions of \( 1 \times 1 \), \( 1 \times 3 \), \( 3 \times 1 \), \( 1 \times 5 \), \( 5 \times 1 \) and \( 3 \times 3 \) respectively. \( \mathrm{DConv}_{3 \times 3} \) denotes dilated convolution. \( \mathrm{DFConv}_{3 \times 3} \) denotes deformable convolution.
 \( \mathrm{F} \) denotes the input feature map.
 \( \mathrm{B_1} \), \( \mathrm{B_2} \), and \( \mathrm{B_3} \) denote the output feature maps of the first three branches following deformable and dilated convolution.
 The output of concatenating  \( \mathrm{B_1} \), \( \mathrm{B_2} \), and \( \mathrm{B_3} \) is added to the result of \( \mathrm{Conv}_{1 \times 1} \) on F to obtain the output feature map \( \mathrm{Y} \) of CFE.
This residual connection integrates shallow and deep features, preventing information loss during layer transmission and enriching feature representation.

\subsection{Attention Fusion BiFPN (AFBiFPN)} 
Traditional attention mechanisms reduce computational complexity through fixed or manually designed sparse patterns, but this comes at the cost of sacrificing the ability to capture global information\cite{23}. In contrast, BA dynamically selects the most relevant regions through a dual-layer routing mechanism, achieving content-aware sparse attention. Our proposed AFBiFPN fully leverages the capability of BiFPN to assign dynamic weights to input nodes, allowing the network to automatically learn the relative importance of features at different hierarchy levels. By inserting BA at the fusion nodes, we leverage region-level routing and sparse attention to adaptively strengthen cross-scale interactions, yielding an attention-fused AFBiFPN. BA unfolds in three phases: (1) Region partition and projection, (2) Region-to-region attention, and (3) Token-to-token attention.

Region partition and projection: Given the input feature map $\mathrm {F \in \mathbb{R}^{H \times W \times C}}$. Partition F into \(\mathrm S \times \mathrm S \) regions to obtain feature map $\mathrm {F_{r} \in \mathbb{R}^{S^2 \times \left( \frac{H}{S} \times \frac{W}{S} \times C \right)}}$. Perform linear projections on $\mathrm{F}_r$ to respectively obtain the query ($\mathrm{Q}_r$), key ($\mathrm{K}_r$), and value ($\mathrm{V}_r$) tensors, $\mathrm {Q_r,K_r,V_r \in \mathbb{R}^{S^2 \times \left( \frac{H}{S} \times \frac{W}{S} \times C \right)}}$.

Region-to-region attention: Construct an affinity graph between regions to determine the semantic relationships between each region and the others. The specific steps are divided into the following three steps. Firstly, extract region-level queries and keys from each region using average pooling operations:
\begin{equation}
Q_{r-m}^{(i)} = \text{mean}(Q_r^{(i)}), \quad K_{r-m}^{(i)} = \text{mean}(K_r^{(i)}).
\end{equation}
Secondly, compute the affinity graph of \(Q_{r-m}\) and \(K_{r-m}\), retain the top-K connections for each region to form the index matrix \(I_{r-m} \):
\begin{equation}
I_{r-m} = \text{TopKIndex}(Q_{r-m}K_{r-m}^{T}).
\end{equation}

Token-to-Token Attention: To ensure that query vectors focus only on the most relevant information inside the current region $r$ we first use the index set ${I_{r-m}}$ to gather the key and value tensors from the neighboring regions $K$ with the highest affinity to $r$. These tensors are concatenated in the order of the neighboring regions. For the $i$ attention head, the aggregated keys and values are as Eq. \eqref{eq:7} \eqref{eq:8}.
\begin{equation}
K_{\text{g}}^{(i)} = \left[ K^{(I_{r-m}(i,1))}, K^{(I_{r-m}(i,2))}, \dots, K^{(I_{r-m}(i,K))} \right]
\label{eq:7}
\end{equation}
\begin{equation}
V_{\text{g}}^{(i)} = \left[ V^{(I_{r-m}(i,1))}, V^{(I_{r-m}(i,2))}, \dots, V^{(I_{r-m}(i,K))} \right]
\label{eq:8}
\end{equation}
As in Eq. \eqref{eq:9}. the query vector $q_j$ is dot-multiplied with the aggregated keys $K_g^{(i)}$, scaled by $\sqrt{d_k}$ and normalized via soft-max to yield the attention weights $a_j^{(i)}$, which are then applied to the aggregated values $V_g^{(i)}$ to produce the output $o_j^{(i)}$. 
\begin{equation}
\alpha_j^{(i)} = \mathrm{softmax} \left( \frac{q_j {K_g^{(i)}}^\top}{\sqrt{d_k}} \right), \quad o_j^{(i)} = \alpha_j^{(i)} V_g^{(i)}
\label{eq:9}
\end{equation}

As in Eq. \eqref{eq:p4f_bra111}, BA first applies a scaled dot-product Softmax between the query matrix $Q_r$ and the aggregated keys $K_g$ to obtain region-level attention weights, uses these weights to aggregate the values $V_g$, and finally adds a local-context embedding $LCE(V_r)$ to produce the fused feature $BA(F)$.
\begin{equation}
\mathrm{BA(F)}
=
\operatorname{Softmax}\!\Bigl(
      \tfrac{\mathrm{Q}_{r}\,\mathrm{K}_{g}^{\top}}{\sqrt{C}}
\Bigr)\,
\mathrm{V}_{g}+LCE(V_r)
\;
\label{eq:p4f_bra111}
\end{equation}

As in Eq. \eqref{eq:12}, \eqref{eq:14} represent the top-down, attention fusion features, whereas Eq. \eqref{eq:11}, \eqref{eq:13}, \eqref{eq:15} and \eqref{eq:16}  correspond to the multi-scale outputs produced by the AFBiFPN. The operator $\operatorname{Resize}(\cdot)$ denotes either up-sampling or down-sampling, and each learnable weight $w_{ij}$ is indexed by the output layer $i$ and the input branch $j$. $\operatorname{BA}(\cdot)$ indicates the feature refined by BA, and $\epsilon$ is a small constant added to prevent numerical instability.
\begin{equation}
P_{5}^O = \frac{w_{51} \cdot P_{5}^I + w_{52} \cdot \text{Resize}(P_{4}^O)}{w_{51} + w_{52} + \epsilon}
\label{eq:11}
\end{equation}
\begin{equation}
P_{4}^F = \frac{w_{41} \cdot P_{4}^I + w_{42} \cdot \text{Resize}(P_{5}^I)}{w_{41} + w_{42} + \epsilon}
\label{eq:12}
\end{equation}
\begin{equation}
P_{4}^O = \frac{w_{43} \cdot P_{4}^I + w_{44}\cdot BA(P_{4}^F) +w_{45} \cdot \text{Resize}(P_{3}^O)}{w_{43} + w_{44} + w_{45} +\epsilon}
\label{eq:13}
\end{equation}
\begin{equation}
P_{3}^F = \frac{w_{31} \cdot P_{3}^I + w_{32} \cdot \text{Resize}(BA(P_{4}^F))}{w_{31} + w_{32} + \epsilon}
\label{eq:14}
\end{equation}
\begin{equation}
P_{3}^O = \frac{w_{33} \cdot P_{3}^I + w_{34}\cdot BA(P_{3}^F) +w_{35} \cdot \text{Resize}(P_{2}^O)}{w_{33} + w_{34} + w_{35} +\epsilon}
\label{eq:15}
\end{equation}
\begin{equation}
P_{2}^O = \frac{w_{21} \cdot P_{2}^I + w_{22} \cdot \text{Resize}(BA(P_{3}^F))}{w_{21} + w_{22} + \epsilon}
\label{eq:16}
\end{equation}

\section{Experiment}

\subsection{The datasets} 
The SSDD~\cite{24} is an SAR ship detection benchmark containing 1,560 TerraSAR-X, Sentinel-1, and Gaofen-3 images with 3,946 labelled ships. It covers resolutions from 1–25 m, dual-pol (HH/HV) modes, and scenes from near-shore to polar seas. The targets span $15\times15$ to $300\times300$ px, with metadata on incidence angles (20° to 50°) and noise levels.

\subsection{Experimental Results and Analysis} 
To validate the performance of different detectors on small, medium, and large ships in the SSDD, the following experiments are conducted in this section. A total of 25 detectors are selected for comparison, as shown in Table~\ref{tab:1}. Ablation experiments for C-AFBiFPN are presented in Table~\ref{tab:2}. The evaluation metrics used include AP (Average Precision), AP$_S$(AP for small ships), AP$_M$(AP for medium ships), and AP$_L$(AP for large ships).

\begin{table}[ht]
\centering
\renewcommand{\arraystretch}{1.5}  
\caption{The Performance Of Different Detectors On the SSDD.}
\label{tab:1}
\resizebox{\columnwidth}{!}{%
\begin{tabular}{cllrrrr}
\toprule
No. & Detector & Backbone & AP   & AP\textsubscript{S} & AP\textsubscript{M} & AP\textsubscript{L} \\
\midrule
1   & ATSS+FPN             & ResNet50 & 0.573 & 0.540 & 0.661 & 0.448 \\
2   & Cascade RCNN+FPN     & ResNet50 & 0.604 & 0.561 & 0.673 & 0.641 \\
3   & Doublehead Faster R-CNN+FPN & ResNet50 & 0.593 & 0.558 & 0.662 & 0.509 \\
4   & Faster R-CNN+FPN     & ResNet50 & 0.590 & 0.558 & 0.656 & 0.533 \\
5   & Mask R-CNN+FPN       & ResNet50 & 0.594 & 0.557 & 0.657 & 0.588 \\
6   & Cascade Mask R-CNN+FPN & ResNet50 & 0.611 & 0.567 & 0.688 & 0.680 \\
7   & FCOS+FPN             & ResNet50 & 0.402 & 0.417 & 0.415 & 0.258 \\
8   & Fovea Align+FPN      & ResNet50 & 0.574 & 0.553 & 0.638 & 0.452 \\
9   & Faster R-CNN GA      & ResNet50 & 0.595 & 0.555 & 0.670 & 0.579\\
10  & RetinaNet GA+FPN     & ResNet50 & 0.494 & 0.462 & 0.554 & 0.444 \\
11  & Guided Anchoring+FPN & ResNet50 & 0.382 & 0.395 & 0.425 & 0.285 \\
12  & Grid R-CNN+FPN       & ResNet50 & 0.531 & 0.505 & 0.588 & 0.454 \\
13  & Hybrid Task Cascade+FPN & ResNet50 & 0.603 & 0.566 & 0.675 & 0.579 \\
14  & Libra R-CNN+FPN      & ResNet50 & 0.595 & 0.558 & 0.666 & 0.532 \\
15  & Reppoints Moment+FPN & ResNet50 & 0.536 & 0.521 & 0.576 & 0.484 \\
16  & RetinaNet Free Anchor+FPN & ResNet50 & 0.558 & 0.535 & 0.625 & 0.465 \\
17  & RetinaNet GHM+FPN    & ResNet50 & 0.568 & 0.553 & 0.630 & 0.624 \\
18  & RetinaNet+FPN        & ResNet50 & 0.557 & 0.521 & 0.632 & 0.499 \\
19  & Faster R-CNN+BiFPN   & ResNet50 & 0.651 & 0.666 & 0.610 & 0.653 \\
20  & YOLO X+FBFPN         & DarkNet-cca & 0.623 & 0.551 & 0.675 & 0.650 \\
21  & YOLO V8+OD-FPN       & DarkNet-53  & 0.646 & 0.597 & 0.729 & 0.652       \\
22  & SSD-300              & VGG-16   & 0.536 & 0.501 & 0.604 & 0.556 \\
23  & SSD-512              & VGG-16   & 0.559 & 0.523 & 0.618 & 0.635 \\
24  & YOLO X               & DarkNet-53 & 0.606 & 0.544 & 0.671 & 0.635 \\
25  & YOLO V8              & DarkNet-53 & 0.629 & 0.577 & 0.726 & 0.650 \\
\textbf{26}  & \textbf{Ours}                 & \textbf{ResNet50} & \textbf{0.702} & \textbf{0.698} & \textbf{0.733} & \textbf{0.701} \\
\bottomrule
\end{tabular}%
}
\end{table}

\begin{table}[htbp]
\centering
\caption{Ablation Experiments on the SSDD dataset.}
\label{tab:2}
\begin{tabular}{ccccccc}
\toprule
BiFPN & CFE & Attention Fusion & AP   & AP\textsubscript{S} & AP\textsubscript{M} & AP\textsubscript{L} \\
\midrule
\checkmark & \texttimes & \texttimes & 0.651 & 0.666 & 0.610 & 0.653 \\
\checkmark & \checkmark & \texttimes & 0.661 & 0.667 & 0.670 & 0.663 \\
\checkmark & \texttimes & \checkmark & 0.690 & 0.698 & 0.703 & 0.670 \\
\textbf{\checkmark} & \textbf{\checkmark} & \textbf{\checkmark} & \textbf{0.702} & \textbf{0.698} & \textbf{0.733} & \textbf{0.701} \\
\bottomrule
\end{tabular}
\end{table}

\begin{figure}[htbp]
    \centering
    \includegraphics[width=0.5\textwidth, height=0.4\textheight]{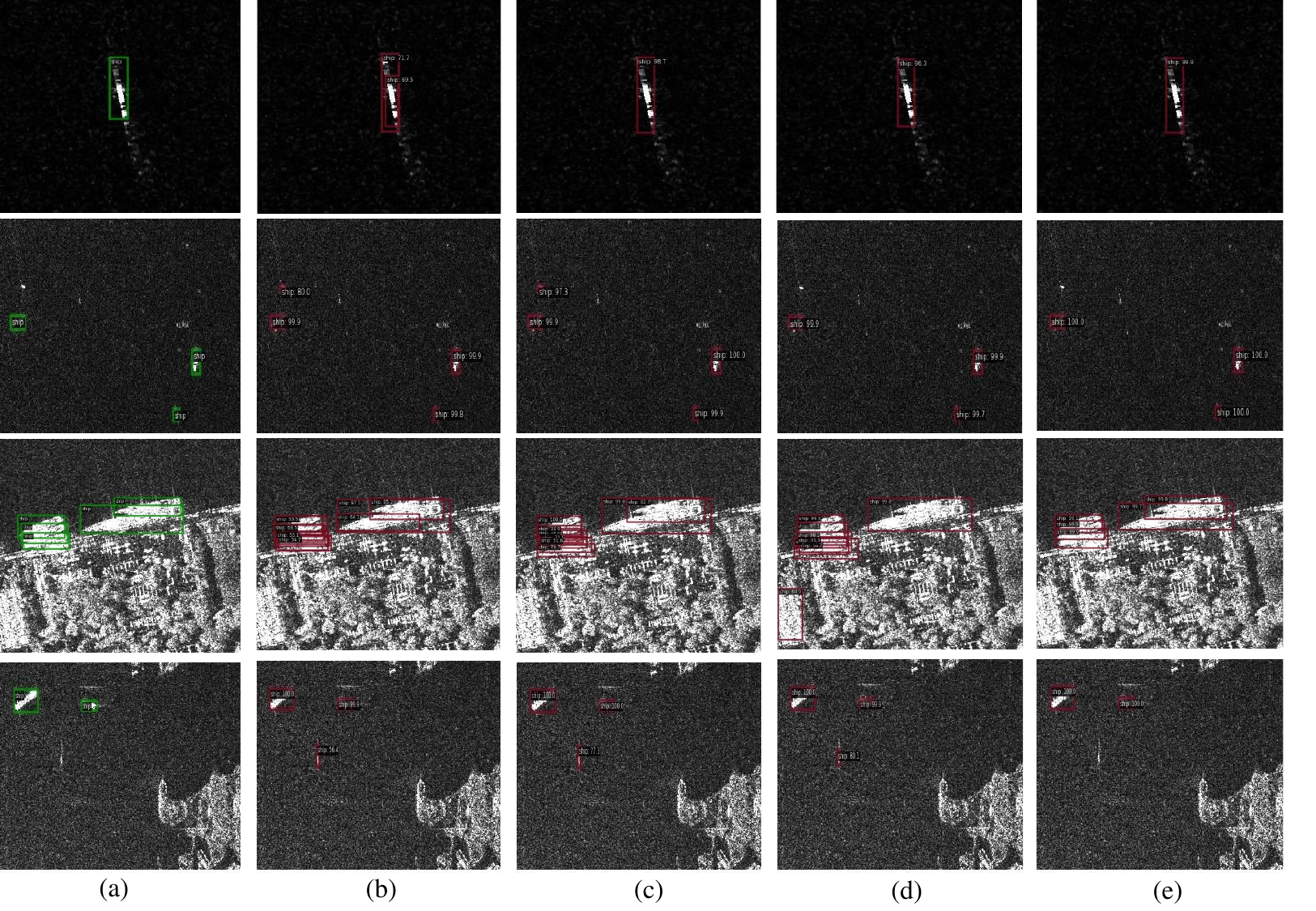} % 
    \caption{Comparative Chart of Ablation Experiments on the SSDD. (a) Ground truth. (b) BiFPN. (c) BiFPN + CFE. (d) AFBiFPN. (e) C-AFBiFPN.}
    \label{fig:4}
\end{figure}

In Table~\ref{tab:1}, we select 25 advanced detectors as benchmarks, including single-stage and two-stage methods. It is evident that our detector outperforms all 25 detectors in all evaluation metrics. Specifically, it achieves an improvement of approximately 7.2\% over the second best in AP, around 4.6\% in AP$_S$, about 0.5\% in AP$_M$, and roughly 3\% in AP$_L$. Specifically, the proposed C-AFBiFPN feature pyramid framework in this paper adopts the two-stage object detector Faster R-CNN as the overall framework, fully leveraging the high accuracy of two-stage object detectors. Compared to the original Faster R-CNN + FPN object detector, Faster R-CNN + C-AFBiFPN shows significant improvements, with 19\% higher AP, 24\% higher AP$_S$, 11.7\% higher AP$_M$, and 31\% higher AP$_L$. These improvements are attributed to the C-AFBiFPN, which enhances feature maps after the out of the backbone network by applying CFE. This improves region awareness, improves weak feature representation for small targets, and suppresses background noise confusion. Furthermore, during the top-down fusion process in FPN, BA utilizes a dual-layer routing mechanism to independently select the most relevant cross-layer regions at each intermediate layer. The region-level routing of BA prioritizes areas with textures similar to ships, making feature fusion more focused on relevant targets.

In Table~\ref{tab:2}, we perform ablation experiments to analyze the performance of the proposed method. When neither CFE nor attention fusion was included, the detector performance was significantly constrained.
After incorporating the CFE, feature enhancement was applied following the backbone network output, allowing the extracted feature maps to better retain contextual information, resulting in improvements across all evaluation metrics. Subsequently, by removing the CFE and retaining only the attention fusion, the feature maps were refined through attention fusion, strengthening critical features at specific scales, which led to notable improvements in all evaluation metrics. Finally, when the CFE was added after the backbone network, the proposed C-AFBiFPN outperformed the original BiFPN in all metrics by a substantial margin.

In Fig.~\ref{fig:4}, Comparative Chart of Ablation Experiments in the SSDD, a comparison is made between BiFPN, BiFPN+CFE, AFBiFPN, and the proposed C-AFBiFPN method in the SSDD for nearshore and offshore scenarios. The BiFPN model exhibits detection errors under different conditions both in nearshore and offshore settings. In contrast, BiFPN+CFE and AFBiFPN provide more accurate detections than BiFPN in both nearshore and offshore scenarios by either obtaining richer semantic information after the backbone network or by refining the fused feature maps to enhance key features at specific scales. Finally, the proposed C-AFBiFPN model demonstrates superior detection performance in multi-scale ship detection scenarios by suppressing false alarms, enhancing features from the backbone network, and strengthening key features at specific scales.

\section{Conclusion}
This paper proposes the novel feature fusion framework C-AFBiFPN that integrates convolutional feature enhancement and attention fusion for multi-scale ship detection in SAR images. 
Firstly, this study identifies the limited feature extraction capability of existing backbone networks, which often leads to missing critical features of ships. To address this issue, a CFE module is designed following the backbone to enrich feature representation and expand the receptive field, thereby enhancing detection performance.
Next, to highlight important information within the fused feature maps while suppressing background interference and enhancing key features at specific scales, this paper proposes AFBiFPN. This allows for the extraction of prominent ship features through the attention mechanism after multi-level feature fusion, effectively reducing background noise and irrelevant information.
Overall, the proposed C-AFBiFPN, as a novel feature fusion framework, effectively extracts the fused feature maps required for multi-scale ship detection, achieving high-precision ship detection across various SAR image scenarios. Experimental results demonstrate that the method accurately detects multi-scale ships in different environments. Furthermore, compared to other advanced ship detection methods, the proposed approach shows superior performance.


% \newpage

 

% \end{IEEEbiography}



% \vfill

\end{document}